\documentclass[10pt, conference]{IEEEtran}
\IEEEoverridecommandlockouts
\usepackage{cite}
\usepackage{amsmath,amssymb,amsfonts}
\usepackage{graphicx}
\usepackage{textcomp}
\usepackage{xcolor}
\def\BibTeX{{\rm B\kern-.05em{\sc i\kern-.025em b}\kern-.08em
    T\kern-.1667em\lower.7ex\hbox{E}\kern-.125emX}}

\usepackage{enumerate}
\usepackage{algorithmicx,algorithm}
\usepackage{multirow}
\usepackage[noend]{algpseudocode}
\usepackage{subfigure}
\usepackage{url}

\graphicspath{{figure1/}}

\usepackage{fancyhdr}

\begin{document}

\title{HADFL: Heterogeneity-aware Decentralized Federated Learning Framework\\
\thanks{* Corresponding author.}
}

\author{
Jing Cao\textsuperscript{1}, Zirui Lian\textsuperscript{1}, Weihong Liu\textsuperscript{1}, Zongwei Zhu\textsuperscript{*1}, Cheng Ji\textsuperscript{2} \\
\IEEEauthorblockA{\textit{\textsuperscript{1} University of Science and Technology of China, China}}
\IEEEauthorblockA{\textit{\textsuperscript{2} Nanjing University of Science and Technology, China}}
\{congjia, ustclzr, lwh2017\}@mail.ustc.edu.cn, zzw1988@ustc.edu.cn, cheng.ji@njust.edu.cn
}

\maketitle
\thispagestyle{fancy}
\fancyhead{}
\lhead{}
\lfoot{978-1-6654-3274-0/21/\$31.00 \copyright 2021 IEEE}
\cfoot{}
\rfoot{}

\begin{abstract}
Federated learning (FL) supports training models on geographically distributed devices. However, traditional FL systems adopt a centralized synchronous strategy, putting high communication pressure and model generalization challenge. Existing optimizations on FL either fail to speedup training on heterogeneous devices or suffer from poor communication efficiency. In this paper, we propose HADFL, a framework that supports decentralized asynchronous training on heterogeneous devices. The devices train model locally with heterogeneity-aware local steps using local data. In each aggregation cycle, they are selected based on probability to perform model synchronization and aggregation. Compared with the traditional FL system, HADFL can relieve the central server’s communication pressure, efficiently utilize heterogeneous computing power, and can achieve a maximum speedup of 3.15x than decentralized-FedAvg and 4.68x than Pytorch distributed training scheme, respectively, with almost no loss of convergence accuracy.
\end{abstract}

\begin{IEEEkeywords}
Distributed Training, Machine Learning, Federated Learning, Heterogeneous Computing
\end{IEEEkeywords}

\section{Introduction}
Traditional Artificial Intelligence (AI) applications, for example, medical image recognition models, are trained by third-party organizations using data collected from medical centers, which requires high computing power. Besides, due to the privacy characteristics of medical images, the data available to them is often very limited and outdated.

One alternative method is to leave the privacy-sensitive data in local devices or data centers, train the model locally, and then transmit only the privacy-insensitive model parameter to perform model aggregation. Federated Learning (FL) \cite{FL} can address the aforementioned communication pressure. In FL system, selected active devices calculate multiple iterations (i.e. local steps) locally based on the local data. Then, they synchronously transmit the model parameters to the central parameter server to perform model aggregation using Federated Average (FedAvg) algorithm \cite{FedAvg}.

However, there are still three challenges in federated learning. 
1) The system configuration of different devices may differ due to variability in CPU, GPU, memory, and so on. 
The unbalanced computing power of devices can exacerbate the straggler problems \cite{dean2013tail} 
 and cause some nodes to fall behind seriously. 
In synchronous parameter iteration strategies as FedAvg adopts, nodes with slow calculation speed will drag down the global iteration pace. 
2) Although the FL framework reduces communication frequency during training, the communication volume is still very huge. The centralized model aggregation strategy of FedAvg can put great communication and computation pressure on the central server, leading to poor scalability and communicational bottleneck. 
3) The geographic distribution of devices tends to be extensive, which brings high communication unreliability. 
If the system cannot handle the suddenly disconnected device well, its performance will suffer a great loss.

There are many efforts that seek to optimize the FL system. 
In order to solve the impact of inconsistent calculation pace on synchronous FedAvg, some optimizations aim at asynchronous model aggregation \cite{sprague2018asynchronous}. However, parameters on laggard nodes are stale and can bring incorrect convergence or increased iterations \cite{SGDyu}. Some research \cite{xie2019asynchronous}\cite{lu2020privacy} conduct weighted model aggregation to reduce the impact of straggler devices by assigning lower weight to devices with stale parameters. The weight of too stale parameters can be too low, resulting in almost no contribution to the model but the wasted communication and computation time. What's more, they all adopt a centralized model synchronization and aggregation method, which can put great communication pressure when there are massive devices. In terms of decentralized FL, gossip communication \cite{hu2019decentralized}\cite{jiang2020decentralised}\cite{lalitha2018fully} can achieve fully decentralized design with no additional system management overhead. However, they all assume that the devices are homogeneous, and aggregate model synchronously, which is not suitable for training model on heterogeneous devices. 

In this paper, we focus on solving the impact of the heterogeneous device computing power on traditional centralized FL systems and propose a heterogeneity-aware decentralized federated learning framework (HADFL). HADFL supports running different local steps asynchronously according to devices' computing power. It adopts a version-sensitive probabilistic partial model aggregation scheme to reduce the impacts of straggler devices on model convergence. What's more, it adopts a decentralized point-to-point communication method, which can eliminate the communication pressure of the central server without increasing the overall communication volume. To the best of our knowledge, this study is the first that fully considers gossip-based decentralized federated learning on heterogeneous devices. Our experiments show that it can achieve a maximum speedup of 3.15x than decentralized-FedAvg \cite{hegedHus2019decentralized} and 4.68x than Pytorch distributed training scheme \cite{horovod}, respectively, with almost no loss of convergence accuracy. The main contributions of this paper are as follows:

\begin{figure}[t] \centering
\includegraphics[width=0.46\textwidth]{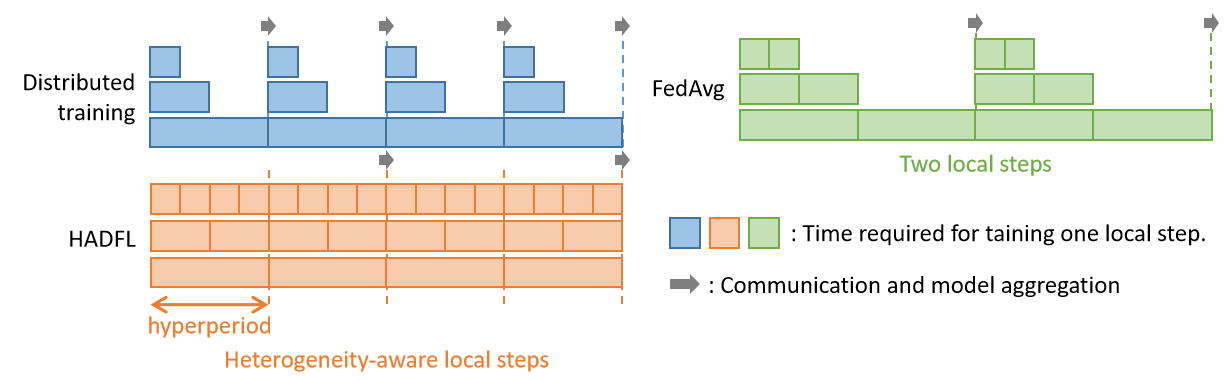}
\caption{The comparison of distributed training, FedAvg and HADFL. There are three devices, and their computing power ratio is 4:2:1.} 
\label{COM}
\end{figure}

\begin{itemize}
\item We propose a heterogeneity-aware asynchronous local training algorithm, which allows heterogeneous devices to run different local steps before model aggregation. A dynamic prediction function is used to predict the parameter versions according to historical operating information, to have good guidance during long-term operation.
\item A decentralized model aggregation strategy is adopted. Devices communicate with each other to transfer model parameters. During model aggregation, to reduce the negative impacts of straggler devices, we propose a probability-based selection method, which not only does not waste the efforts of straggler devices but also can utilize the noise brought by them for faster training.
\item The HADFL framework fully considers the unreliability of the network connection during operation and adopts a fault-tolerant parameter synchronization scheme.
\end{itemize}

\section{Background and Motivation}
\subsection{Model Training}
The model training process can be divided into three phases:
\begin{enumerate}[1.]
\item \emph{Forward propagation}. Calculate output according to the current model parameters using a batch of data.
\item \emph{Backward propagation}. Calculate the $loss$ between the calculated output and the expected output, and calculate gradients of the $loss$ to each model parameter.
\item \emph{Model update}. Update model parameters using the gradients, fetch another batch of data, and repeat the above steps.
\end{enumerate}

We define it an \emph{iteration} to process a $batch\_size$ of data, and an \emph{epoch} to process all samples in the training data set once, which typically contains several \emph{iterations}. What's more, the training typically requires multiple \emph{epochs}. 

The training purpose is to minimize the loss, i.e. $\min \frac{1}{N} \sum_{i=1}^N f(x_i,w)$, where $w$ is the model parameters, $f(x,w)$ is the loss function, $x_i$ is the i-th train sample, and $N$ is the total number of samples in the training set. If there are $K \geq 1$ devices training cooperatively, the \emph{model update} process can be expressed as
\begin{equation}
w_{(t+1)} = w_{(t)} - lr_{(t)} \begin{matrix}\frac{1}{KB}\sum_{k=1}^{K}\sum_{x_i \in \cal{P}\it{_t}}\end{matrix} \triangledown f(x_i, w_{(t)}) \label{Training}
\end{equation}
where $lr_{(t)}>0$ is the learning rate, $\cal{P}\it{^k}$ is the mini-batch of training data of device k, and $B$ is the batch\_{size}. For simplicity, we assume that each device has the same batch\_{size}.

\begin{algorithm}[t]
\caption{Heterogeneity-aware Local Training.}
\label{alg1} 
\small
{\bf Input:}
 the initial model $w_{(0)}$, the batch\_size $B$, $lr_{(t)}$, training data\\
\hspace*{0.4in} on k-th device $\cal{P}\it{^k}$, $T_{sync}$, local step $E_k$, available devices\\
\hspace*{0.4in} $\{N_{avl}\}$, $Flag^k$, total epochs $T_{total}$ \\
{\bf Output:}
the trained model.
\begin{algorithmic}[1]
\State synchronize the initial models $w_{(0)}^k=w_{(0)}$ for $k \in \{N_{avl}\}$
\State $t_{syn}=0$ 
\For{$t=1$ to $T_{total}$}
\For{\textbf{all} $k \in \{N_{avl}\}$}
\textbf{in parallel}
\For{$e_k=0$ to $E_k$}
\If{$t \geq T_{sync}t_{syn}$} 
\State $e_k=0$
\State $t_{syn}=t_{syn}+1$
\State // partial synchronization
\State $w_{(t+1)} = \begin{matrix}\frac{1}{K} \sum_{k=1}^{K} Flag^k \times w_{(t)+E_k}^k\end{matrix}$
\State // global synchronization
\State send $w_{(t+1)}$ to devices whose $Flag^k=0$
\Else
\State $t=t+1$
\State sample a mini-batch from $\cal{P}\it{^k_{(t)}}$
\State compute the gradient: 
\State $g_{(t)+e_k}^k = \begin{matrix}\frac{1}{B} \sum_{x_i \in \cal{P}\it{^k_{(t)+e_k}}}\end{matrix} \triangledown f(x_i, w_{(t)+e_k})$
\State update the local model: 
\State $w_{(t)+e_k+1}^k=w_{(t)+e_k}^k - lr_{(t)}g_{(t)+e_k-1}^k$ 
\EndIf 
\EndFor 
\EndFor 
\EndFor
\end{algorithmic} 
\end{algorithm}

\subsection{Federated Learning}
In FL, devices transmit model parameters to the server every $E$ local steps. Then, the server executes model aggregation. The local steps of different devices are the same. 

Assuming that $\cal{P}\it{^k}$ contains $n_k$ samples, the training purpose becomes 
\begin{equation}
\min \begin{matrix}\sum_{k=1}^{K} \frac{n_k}{N} \sum_{x_i \in \cal{P}\it{^k}} \frac{1}{n_k} f(x_i,w)\end{matrix}
\end{equation}

The training process becomes \cite{lin2018don}:
\begin{equation}
w_{(t)+e+1}^k=w_{(t)+e}^k - lr_{(t)} \begin{matrix}\frac{1}{B} \sum_{x_i \in \cal{P}\it{^k_{(t)+e}}}\end{matrix} \triangledown f(x_i, w_{(t)+e}) \label{Local}
\end{equation}
\begin{equation}
w_{(t+1)} = \begin{matrix}\frac{1}{K} \sum_{k=1}^{K} w_{(t)+E}^k\end{matrix} \label{Global}
\end{equation}
where $w_{(t)+e}^k$ denotes the parameter on device k after $t$ communication rounds and $e$ local steps. Formula \eqref{Local} represents the local update on local devices and formula \eqref{Global} represents the global model aggregation on central server. Since the gradient size equals to model parameter size, the server needs to communicate data of $2 \times M \times K \times epoch\_num / E$ size during training, where $M$ is the model size. The total communication volume of devices is $2 \times K \times M$.

\subsection{Heterogeneity-aware Asynchronous Federated Learning}
FL assumes that the devices are homogeneous. When applied to heterogeneous devices, fast devices need to wait for slow devices, causing wasted computing power of fast devices.

To solve this problem, in this paper, we propose a heterogeneity-aware asynchronous federated learning mechanism. Take an example, as shown in Fig.\ref{COM}, the devices compute different steps locally during the hyperperiod (the least common multiple of the training time each epoch of the devices), and only aggregate model every $T_{sync}$, which is a positive integer, multiples of the hyperperiod.

\begin{figure*}[htb] \centering
\includegraphics[width=0.87\textwidth]{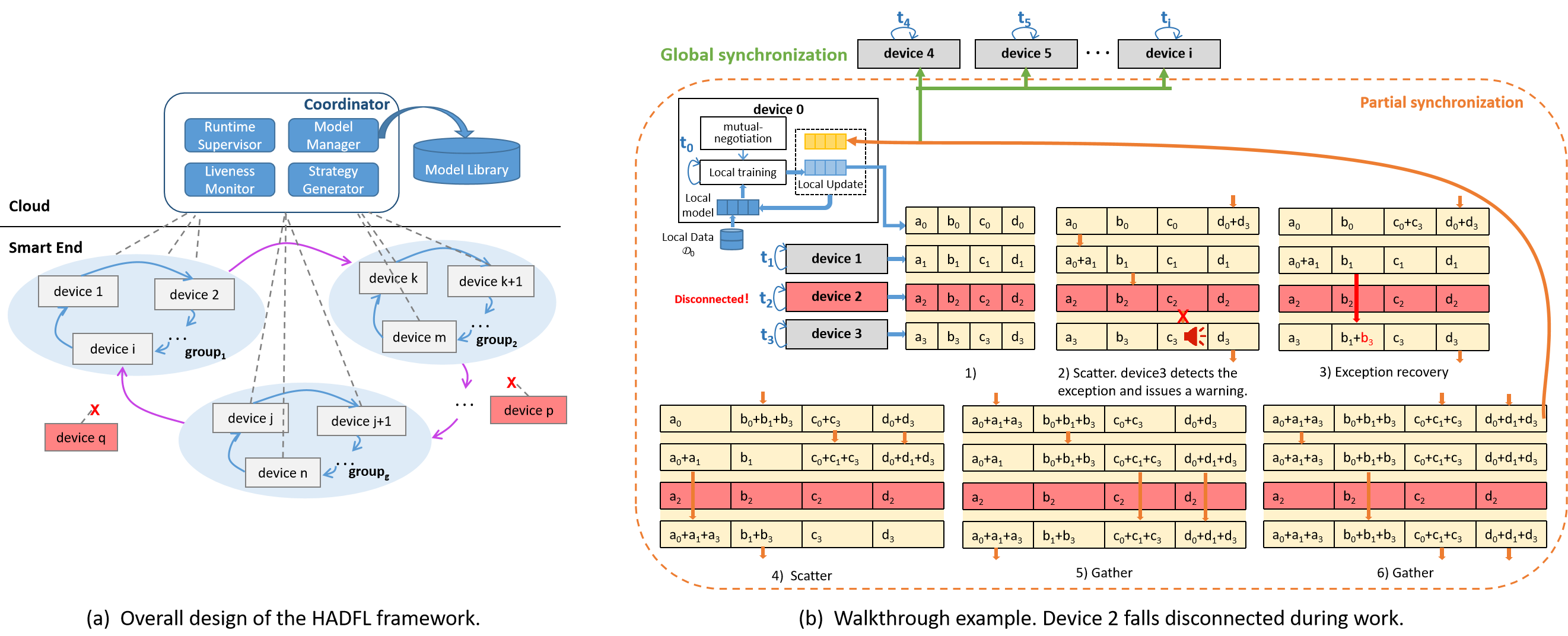}
\caption{The HADFL framework.} 
\label{HADFL}
\end{figure*}

As shown in Algorithm \ref{alg1}, the training process and the local update on local devices are the same as FL, but the global model aggregation becomes:
\begin{equation}
w_{(t+1)} = \begin{matrix}\frac{1}{K} \sum_{k=1}^{K} Flag^k \times w_{(t)+E_k}^k\end{matrix}\label{Global_asyn}
\end{equation}
where $E_k$ is the number of local steps of k-th device. $Flag^k = 1$ if the k-th device is selected for model aggregation, and $Flag^k = 0$ if not.

\section{HADFL Framework}
This section introduces HADFL, a framework that supports decentralized model training on heterogeneous devices. 
It is organized as follows: Section \ref{Overall} shows the overall design of HADFL and the function of each component. Section \ref{predict} introduces the runtime parameter version prediction module. Section \ref{strategy} shows how the heterogeneity-aware local training strategy is generated. Section \ref{aggregation} introduces the partial model aggregation scheme and fault-tolerant strategy.

\subsection{Overall Design} \label{Overall}
As shown in Fig.\ref{HADFL} (a), the HADFL framework consists of a cloud coordinator and several devices. The cloud coordinator performs initial model dispatch, training strategy generation, runtime management, and model backup. It consists of four components: \emph{runtime supervisor}, \emph{liveness monitor}, \emph{strategy generator}, and \emph{model manager}. The devices are responsible for training the model locally, reporting runtime information to the coordinator, and updating the model.

The system workflow is as follows:
\begin{enumerate}[1.]
\item Before the start of each round, the \emph{liveness monitor} module of cloud coordinator first monitors the status of each device and adds the available devices to this round of training.
\item After determining all available devices, \emph{strategy generator} sends training configuration (i.e. the initial model parameters and training hyper-parameters) to devices.
\item Then, each device i enters the \emph{mutual-negotiation} phase and sends its calculation time $T_i$ in this phase to the coordinator, which can reflect its computing power.
\item The \emph{strategy generator} determines the training configuration, including the local step $E_i$, the synchronization period $T_{sync}$ and partial synchronization topology using the distribution of $T_i$, expected parameter version and the probability-based selection function. The design details of \emph{strategy generator} will be introduced in section \ref{strategy}.
\item Each device conducts local training asynchronously according to the training configuration information.
\item Model synchronization. After reaching $T_{sync}$, devices conduct partial model synchronization according to the topology given by the coordinator and broadcast the synchronized model to the other devices in a non-blocking way.
\item Dynamic configuration update. The \emph{runtime supervisor} collects devices' parameter version in each communication round, predicts the parameter version distribution in the next round (the design details will be introduced in section \ref{predict}), and sends it to the \emph{strategy generator} to generate the new training configuration.
\item Repeat the step (4) to (7) until the model converges.
\item Model backup. The \emph{model manager} regularly fetches the latest model and puts it in the database for backup.
\end{enumerate}

\subsection{Runtime information prediction} \label{predict}
During the \emph{mutual-negotiation} phase, the device 1) trains $E_{warm\_up}$ epochs using a small learning rate, which can alleviate the severe fluctuations caused by large loss of the model prediction at the early stage of training
and help to maintain the stability of the model \cite{gotmare2018closer}\cite{he2016deep}, and 2) sends its calculation time in this phase to the coordinator. 

Since the model calculated by each device and the $batch\_size$ used are the same, the calculation time $T_i$ is inversely proportional to the i-th device's computing power. The coordinator then calculates the expected model version 
\begin{equation}
\hat{v_i} = T_{sync}*T_i/E_{warm\_up}
\end{equation}

However, the system may be disturbed during training, causing varying training time. As a result, the expected model version should be updated using historical data dynamically. The \emph{runtime supervisor} collects devices' actual parameter version in each model synchronization round, and predicts the expected model version in the next round using: 
\begin{equation}
\begin{cases}
\hat{v_{i,j+m}} = a_{i,j} + b_{i,j}m &\text{where:} \\
a_{i,j} = 2v_{i,j}^{(1)} - v_{i,j}^{(2)}, \\
b_{i,j} = \frac{\alpha}{1-\alpha}(v_{i,j}^{(1)} - v_{i,j}^{(2)}), \\
v_{i,j}^{(1)} = \alpha v_{i,j} + (1-\alpha)v_{i,j-1}^{(1)}, \\
v_{i,j}^{(2)} = \alpha v_{i,j}^{(1)} + (1-\alpha)v_{i,j-1}^{(2)}
\end{cases}
\end{equation}
in which $v_{i,j}$ is the actual parameter version of the device i in the j-th round, $\hat{v_{i,j+m}}$ is the predicted version in the (j+m)-th round, $v_{i,j}^{(k)}$ is the k-th order exponent of $v_{i,j}$. $0<\alpha<1$ is the smoothing factor, which indicates the weight of $v_i$ during prediction. The larger $\alpha$, the closer the predicted value to $v_i$.

\begin{figure*}[]
	\centering
	\subfigure[Loss vs. epoch on Resnet-18.]{
			\includegraphics[width=0.27\textwidth]{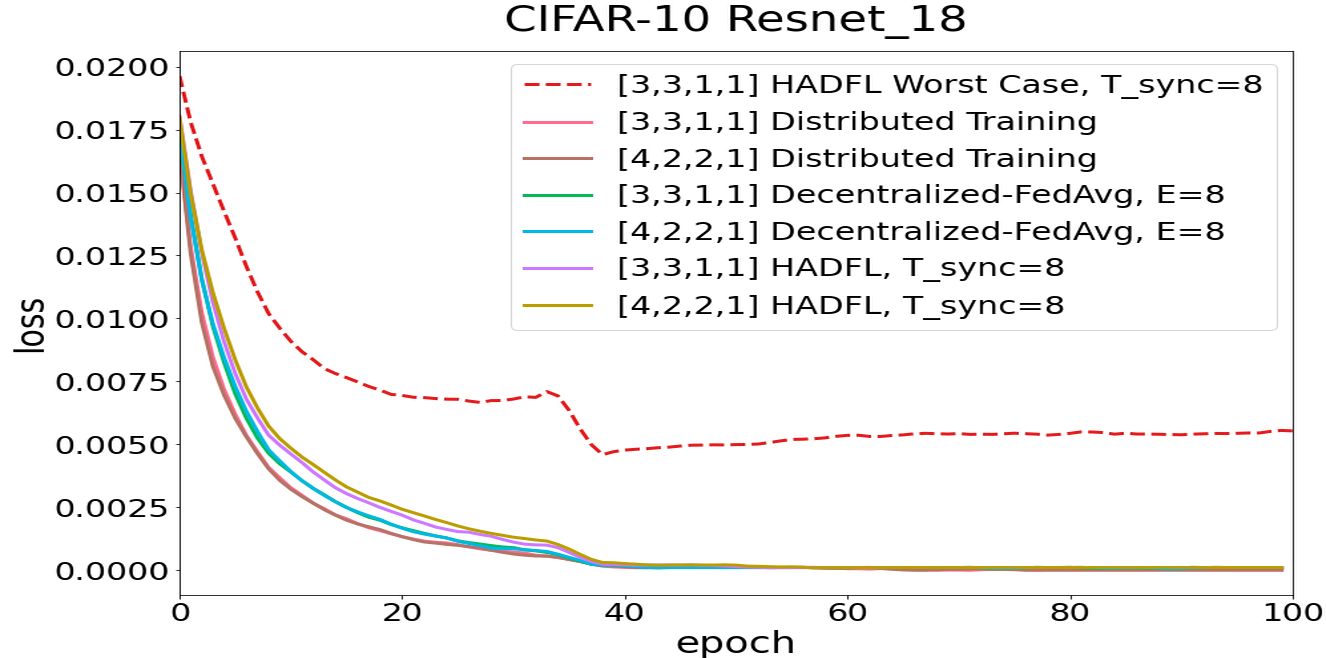}
	}
	\subfigure[Test accuracy vs. epoch on Resnet-18.]{
			\includegraphics[width=0.27\textwidth]{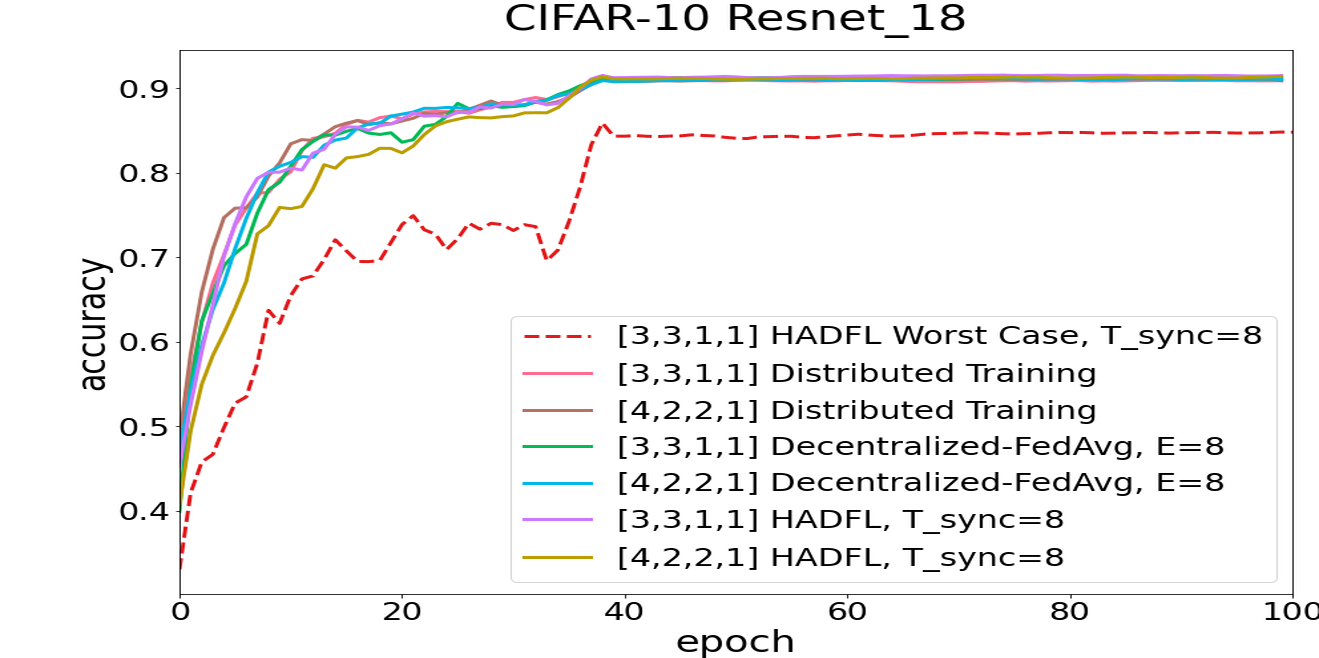}
	}
	\subfigure[Test accuracy vs. time on Resnet-18.]{
			\includegraphics[width=0.27\textwidth]{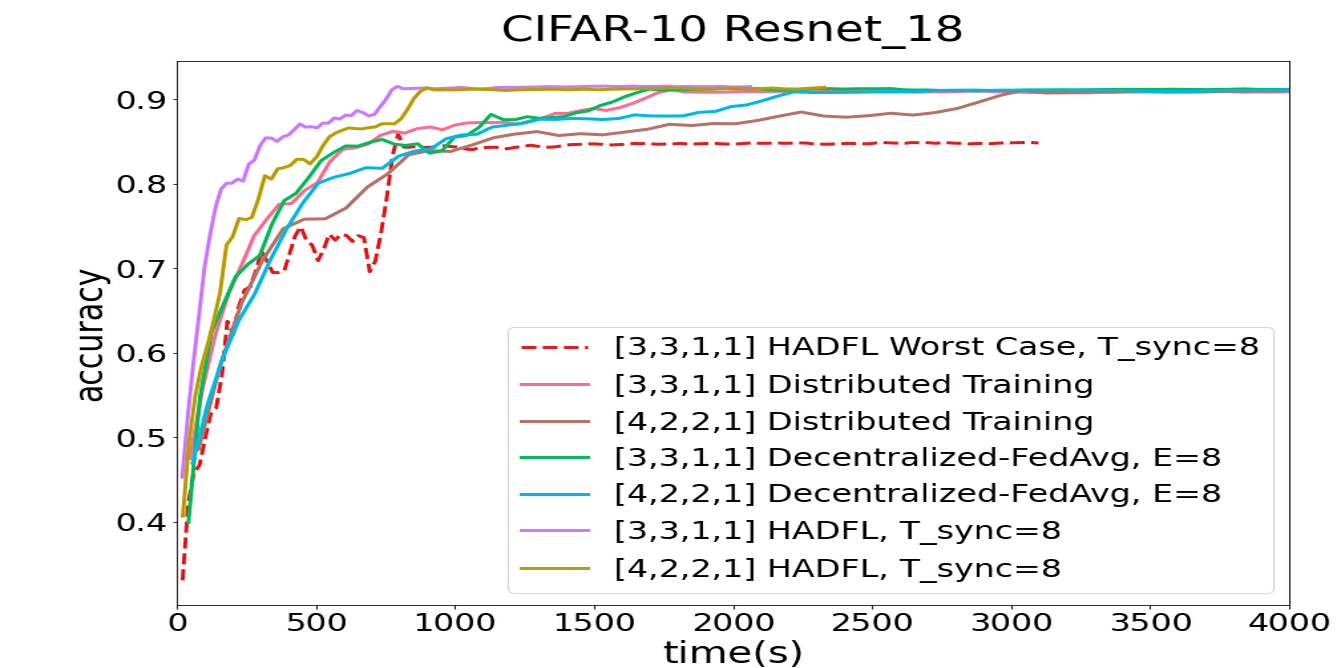}
	}
	\subfigure[Loss vs. epoch on vgg-16.]{
			\includegraphics[width=0.27\textwidth]{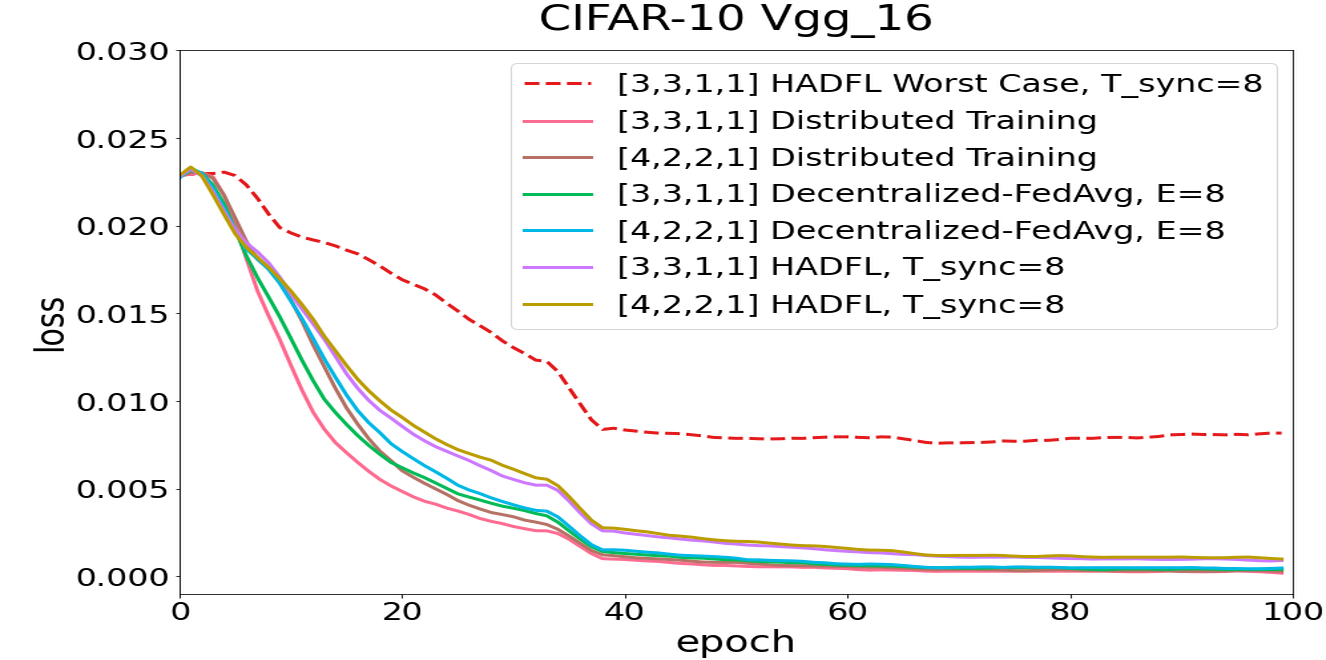}
	}
	\subfigure[Test accuracy vs. epoch on vgg-16.]{
			\includegraphics[width=0.27\textwidth]{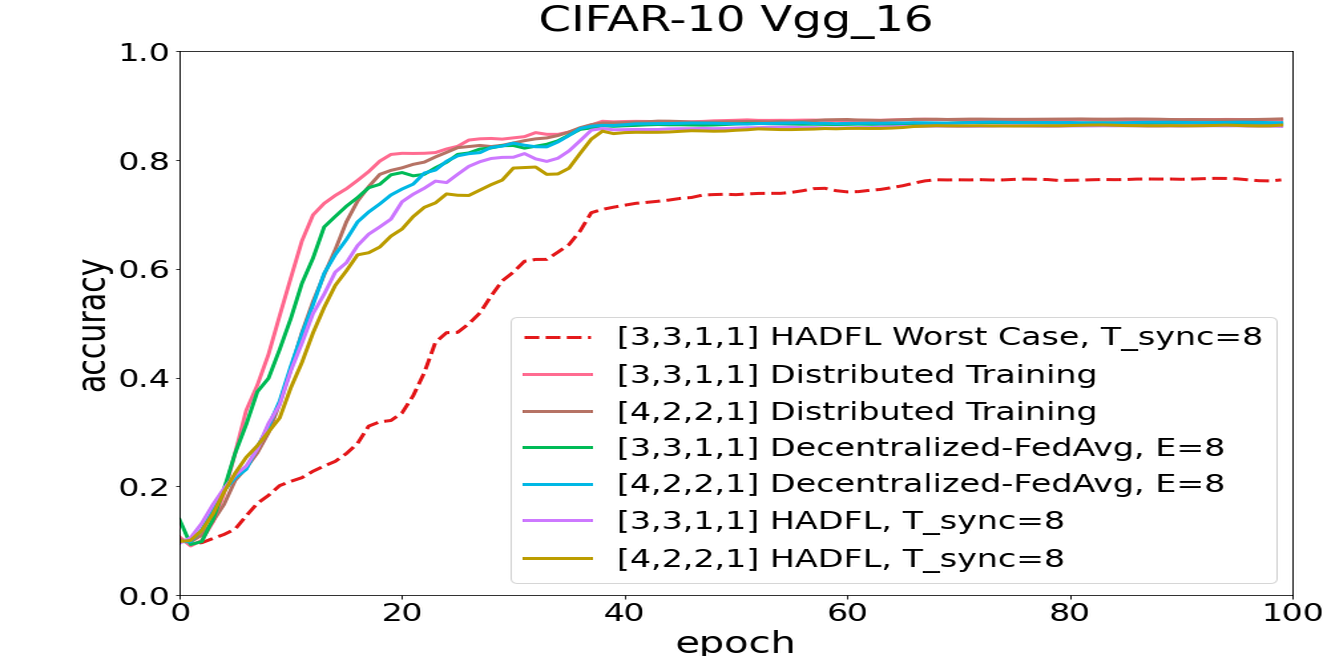}
	}
	\subfigure[Test accuracy vs. time on vgg-16.]{
			\includegraphics[width=0.27\textwidth]{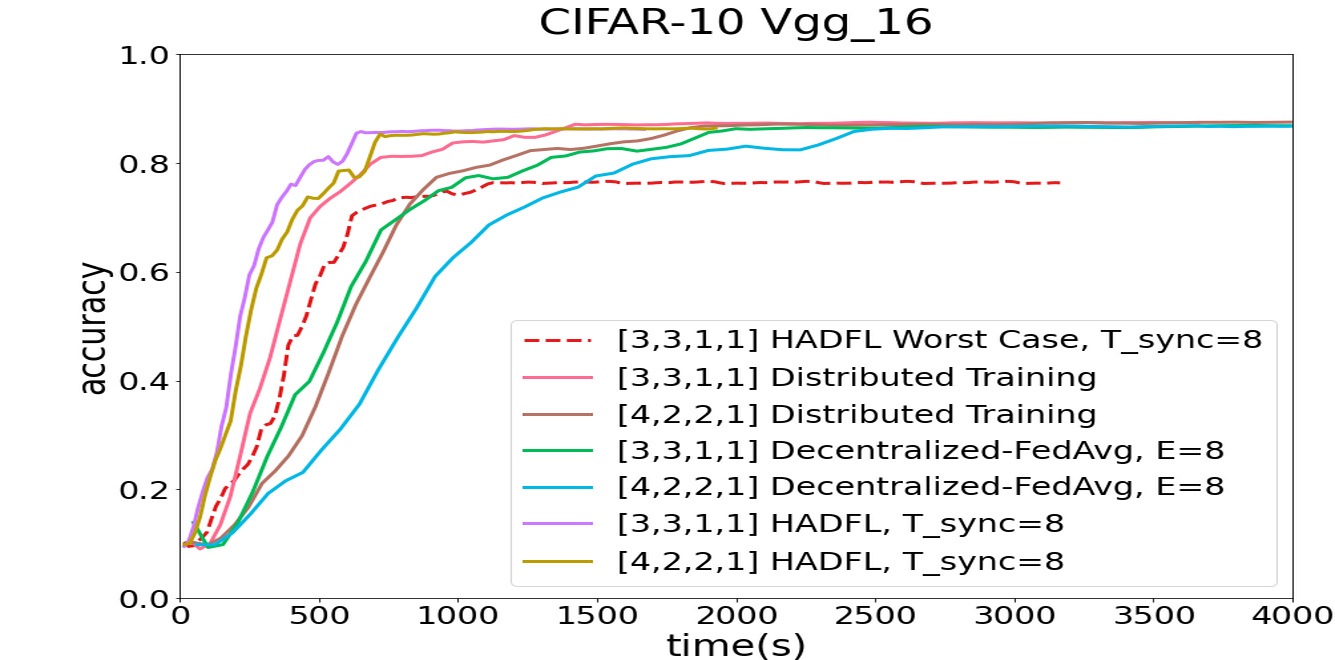}
	}
	\caption{The experimental results.}
	\label{exp}
\end{figure*}

\begin{table*}[ht] \centering
\caption{Time required to reach the maximum test accuracy}  \label{Compare}
\begin{tabular}{|c|c|c|c|c|c|c|c|c|}
  \hline
                  & \multicolumn{2}{c|}{ResNet-18 [3,3,1,1]}     & \multicolumn{2}{c|}{ResNet-18 [4,2,2,1]} & \multicolumn{2}{c|}{vgg-16 [3,3,1,1]} & \multicolumn{2}{c|}{vgg-16 [4,2,2,1]} \\
  \cline{2-9}  
                  &accuracy &time     &accuracy &time    &accuracy &time    &accuracy &time \\
  
  \hline
  Distributed training   &91\% &2431.38 s    &91\% &4076.28 s   &87\% &1349.73 s    &87\% &1791.36 s \\
  Decentralized-FedAvg   &91\% &1699.05 s    &91\% &2747.12 s   &86\% &1952.01 s    &86\% &2424.12 s \\
  HADFL                  &90\% &805.00 s        &91\% &871.50 s     &86\% &794.02 s    &86\% &1324.04 s \\
  \hline
\end{tabular}
\end{table*}

\subsection{Heterogeneity-aware training strategy generation} \label{strategy}
Define the hyperperiod $H_E$ as the least common multiple of the one epoch training time of the devices, i.e. $H_E=LCM_{i=1}^{N_{avl}}(T_i/E_{warm\_up})$. $N_{avl}$ is the total number of devices available. Then, partial aggregation takes place every $T_{sync}$ multiples of $H_E$.

The \emph{strategy generator} uses \eqref{prob}, the probability-based selection function, to determine the probability of each device being selected. $\mu$ is the 3rd quartile of all $v_{i,j}$. Then, it selects $N_{p}$ devices to perform partial synchronization.
\begin{equation}
\begin{cases}
P(i,j)=f(v_{(i,j)})/\sum_{n=1}^{N_{avl}}f(v_{(n,j)}) &\text{where:} \\
f(x)=\int \frac{1}{\sqrt{2\pi}}\exp \left( - \frac{(x-\mu)^2}{2} \right) \label{prob}
\end{cases}
\end{equation}

The probability-based selection function can ensure that the devices with newer parameters (i.e. larger $\hat{v_{i,j}}$) has a higher probability of being selected, thereby reducing the influence of straggler device's parameters on model convergence. However, the straggler devices should not be completely discarded, otherwise, their computing power will be wasted. What's more, their parameters can bring some noise, thereby helping the model to jump out of the local minimum and converges more quickly. In addition, to balance the version differences of all running devices, the devices owning medial versions have a greater probability of being selected, rather than the devices that have the latest parameters. After determining the selected device, the \emph{strategy generator} randomly determines a directed ring as the partial synchronization topology.

If there are too many devices available, in order to facilitate management and avoid possible system errors, the devices can be divided into multiple groups, as shown in Fig.\ref{HADFL} (a). The inter-group synchronization period can be an integer multiple of the intra-group synchronization period. They are performed separately during the training process. The strategy of inter-group synchronization is similar to that of intra-group synchronization, as shown in Fig. \ref{HADFL} (b). 

\subsection{Model aggregation and fault-tolerant} \label{aggregation}
The devices compute gradients and update model parameter asynchronously during their local steps. After reaching the synchronization time $T_{sync}$, as shown in Fig.\ref{HADFL} (b), the selected devices transfer parameters to each other in a gossip-based scatter-gather manner (similar to \cite{horovod}), and perform partial model aggregation and synchronization. Then, a random device in the partial synchronization topology, e.g. device 0 in Fig.\ref{HADFL} (b), transmits the latest model parameters to the unselected $K-N_P$ (typically $\leq K/2$) devices in a non-blocking manner, which will integrate the received model parameters with local parameters and conduct the next round of local training. The total communication volume of devices is $2 \times K \times M$, which is the same as FL.

In order to avoid system errors caused by unstable network connections, we propose a fault-tolerant mechanism. As shown in Fig.\ref{HADFL} (b), for example, device 2 falls disconnected during work, causing its downstream device, device 3, cannot receive parameters in model synchronization. After the pre-specified waiting time, device 3 sends a handshake message to device 2 to confirm its status. After confirmation, it issues a warning to device 1, the upstream of device 2. Then, device 1 will bypass device 2 and communicate directly with device 3.

\section{Experimental Evaluation}
\subsection{Experimental Setup}
\textbf{Testing Platform Setting:} We deploy HADFL framework on four Nvidia Tesla V100 GPUs, which communicate with each other using PCIE Express 3.0 x8. The CUDA version is 10.0.130. We use the $sleep()$ function to simulate different degrees of heterogeneity and use an array to represent the computing power ratio. For example, $[2,1]$ means that the computing power of GPU 0 is twice that of GPU 1.

\textbf{Model and Dataset:} Two CNN model are used as our testing targets, namely, ResNet-18 \cite{he2016deep} and vgg-16 \cite{vgg}. The dataset is CIFAR-10 \cite{CIFAR}, which contains 60K $32 \times 32$ color images. The learning rate is as \cite{he2016deep} adopts in \emph{mutual-negotiation} phase and 0.01 in other phase. The global batch\_size is 256, i.e. the batch\_size on each GPU is $256/4=64$.

\textbf{Comparison benchmark:} To exhibit the effectiveness and superiority of our proposed HADFL framework, we adopt two training schemes for comparison: (1) \emph{Distributed training }\cite{horovod}. We choose the Pytorch distributed training scheme. It uses a decentralized ring all reduce algorithm, and is widely used in distributed training. (2) \emph{Decentralized Federated Average (Decentralized-FedAvg)} \cite{hegedHus2019decentralized}. In Decentralized-FedAvg, devices use a gossip-based method to transmit gradients to peers and merge gradients from peers synchronously.

\subsection{Results}
We run comparative experiments on system with two kind of heterogeneity distribution, $[3,3,1,1]$ and $[4,2,2,1]$. The training data is spilt on four GPUs. We choose two GPUs to perform partial synchronization each time. The experiments are repeated three times. The experimental results are shown in Fig. \ref{exp}. In addition, we record the average time required to reach the maximum test accuracy, as shown in Table \ref{Compare}.

\textbf{The convergence speed:} As shown in Fig. \ref{exp} (c), (f) and Table \ref{Compare}, thanks to the heterogeneity-aware asynchronous strategy, HADFL converges faster than the other two schemes. When training ResNet-18, it achieves 3.02x speedup over distributed training and 2.11x speedup over decentralized-FedAvg in heterogeneous distribution of $[3,3,1,1]$, as well as 4.68x speedup over distributed training and 3.15x speedup over decentralized-FedAvg in heterogeneous distribution of $[4,2,2,1]$, respectively. When training vgg-16, it achieves 1.70x speedup over distributed training and 2.46x speedup over decentralized-FedAvg in heterogeneous distribution of $[3,3,1,1]$, as well as 1.35x speedup over distributed training and 1.83x speedup over decentralized-FedAvg in heterogeneous distribution of $[4,2,2,1]$, respectively. It's worth noting that when training vgg-16 on decentralized-FedAvg, it needs more time to converge than distributed training. This is because the local update is conducted on the local model, which is slightly outdated and can bring loss of accuracy. As a result, it requires more epochs to converge. HADFL also suffers this accuracy loss.

\textbf{The accuracy loss:} As shown in Fig. \ref{exp} (a), (b), (d) and (e), under the same epoch number, the loss of HADFL is a little bit larger than the other schemes, which is caused by our partial synchronization and local update strategy. In $[4,2,2,1]$ heterogeneity distribution, HADFL suffers a slight drop in accuracy every epoch. However, it can also reach almost the same converge test accuracy as the other two schemes. By allowing more GPUs to participate in partial synchronization, the training effect can be better, which is because the waste of efforts on unselected devices is less. What's more, the \emph{mutual-negotiation} phase can make HADFL more stable at the beginning of training, as shown in Fig. \ref{exp} (e) and (f).

\textbf{Upper bound of accuracy loss:} The accuracy loss caused by HADFL has an upper bound. To prove this, we manually specify that during local synchronization, only the two GPUs with the worst computing power are selected each time, and run experiments on GPUs of $[3,3,1,1]$ heterogeneity distribution. As shown in Fig. \ref{exp}, in the worst case, the loss and accuracy fluctuate greatly during the training process, achieving 86\% accuracy on ResNet-18 and 76\% accuracy on vgg-16. This is because only the local data on GPU 2 and GPU 3 are available for model update, and the data on GPU 0 and GPU 1 is wasted for they cannot participate in model aggregation. However, the theoretically probability of this taking place is only $(\frac{1}{8} \times \frac{1}{8})^{epoch_{total}/T_{sync}}$ ($epcoh_{total}$ is the total number of epochs during training), which infinitely approaches 0. 

\section{Related Work}
\subsection{Decentralized Federated Learning}
There is some work using blockchain to design decentralized FL systems \cite{pokhrel2020decentralized}\cite{zhao2020privacy}, however, the management of the blockchain can bring additional delays.
One alternative design is to use gossip communication. 
A. Lalitha et al. \cite{lalitha2018fully} and I. Heged{\H{u}}s et al. \cite{hegedHus2019decentralized} propose a fully decentralized FL scheme in which devices communicate with their neighbours to perform model synchronization. However, this scheme assumes the network is strongly connected, which is not applicable in actual application scenarios with the unstable network connection. 
To solve this problem, \cite{hu2019decentralized} and \cite{jiang2020decentralised} adopt a segmented gossip approach. The model is split into $S$ segmentations, each device is responsible for one segmentation, and sends it to the other $R$ devices. 

Unfortunately, the above work all adopt a synchronized parameter synchronization and aggregation strategy. If applied to heterogeneous systems, devices with slow calculation speed will slow down the training. 

\subsection{Federated Learning on Heterogeneous Devices}
M. R. Sprague et al. \cite{sprague2018asynchronous} propose to let devices pass the parameters to the server for model aggregation immediately after completing the calculation without waiting for slow devices \cite{SGDyu}. However, parameters on straggler devices may be too stale and can bring incorrect convergence or increased iterations. 
W. Wu et al. \cite{wu2019safa} divide devices into three states: latest, deprecated and tolerable according to their model version, and only the latest and deprecated devices are allowed to read new global model from the server. 
In \cite{xie2019asynchronous} and \cite{lu2020privacy}, weighted model aggregation is proposed. Devices with stale parameters are assigned lower weight. However, the weight of too stale parameters can be too low, resulting in almost no contribution to the model but the wasted communication and computation time. 
Y. Chen et al. \cite{chen2019communication} and E. Diao et al. \cite{diao2020heterofl} put different layer structures on heterogeneous devices. This approach relies on communication robustness and can perform poorly in systems with a large amount of devices. 
T. Nishio et al. \cite{nishio2019client} consider using device selection to meet the stale bound, which can cause devices with poor computing power to never be selected.

However, the above work all adopt a centralized model synchronization and aggregation method, which can put great communication pressure when there are massive devices. In this paper, we combine the design ideas of decentralization and asynchronous training and propose a version-based probabilistic device selection scheme. Our framework can alleviate the straggler problem without discarding the efforts of slow devices as well as reduce communication.

\section{Conclusion and future work}
The HADFL framework we propose can support decentralized training on heterogeneous devices efficiently. Our experiments show that it can achieve a maximum speedup of 3.15x than decentralized-FedAvg and 4.68x than Pytorch distributed training scheme, respectively, with almost no loss of convergence accuracy.

In the future, we will deploy the HADFL framework on larger-scale systems, and optimize it by taking into account heterogeneous network bandwidth and data distribution.

\section*{Acknowledgment}
This work was supported by the China Postdoctoral Science Foundation (No. 2020M671637), National Science Youth Fund of Jiangsu Province (No. BK20190224, No. BK20200462), and the Jiangsu Postdoctoral Science Foundation (No. 2019K224).

\bibliographystyle{IEEEtran}
\bibliography{mycite}

\end{document}